%% file: root.tex
\documentclass[letterpaper, 10 pt, conference]{ieeeconf}  

\IEEEoverridecommandlockouts                              

\usepackage{graphicx} 
\usepackage{amssymb}  
\usepackage{amsmath}
\usepackage{color}
\usepackage{float}
\usepackage[ruled,vlined]{algorithm2e}
\usepackage{threeparttable}
\usepackage{multirow}
\usepackage[subrefformat=parens,labelformat=parens]{subfig}
\newtheorem{theorem}{Theorem}[section]
\newtheorem{Problem}[theorem]{Problem}

\title{\LARGE \bf
 Sampling-based path planning under temporal logic constraints with real-time adaptation 
}

\author{Yizhou Chen,  Ruoyu Wang, Xinyi Wang, and  Ben M. Chen
\thanks{	The work  was supported in part by the Research Grants Council of Hong Kong SAR under Grants 14209020 and 14206821,  
   and in part by the Hong Kong Centre for Logistics Robotics.  Authors are with the Chinese University of Hong Kong,
	Shatin, N.T., Hong Kong 999077. (Email: josephchen@link.cuhk.edu.hk;  rywang@link.cuhk.edu.hk; xywangmae@link.cuhk.edu.hk; bmchen@cuhk.edu.hk).
}
}

\graphicspath{{./fig/},{./logo/},{../logo/},{./pic/}}
\begin{document}

	\maketitle
	\thispagestyle{empty}
	\pagestyle{empty}

	\begin{abstract}

    Replanning in temporal logic tasks is extremely difficult during the online execution of robots. This study introduces an effective path planner that computes solutions for temporal logic goals and instantly adapts to non-static and partially unknown environments. Given prior knowledge and a task specification, the planner first identifies an initial feasible solution by growing a  sampling-based search tree.
    While carrying out the computed plan, the robot maintains a solution library to continuously enhance the unfinished part of the plan and store backup plans.  The planner updates  existing plans when meeting unexpected obstacles or recognizing flaws in  prior knowledge. Upon a high-level path is obtained, a trajectory generator tracks the path by dividing it into segments of motion primitives. Our planner is integrated into an autonomous mobile robot system, further deployed on a multicopter with limited onboard processing power.  In simulation and real-world experiments, our planner is demonstrated to swiftly and effectively adjust to environmental uncertainties.

	\end{abstract}

\section{Introduction}
In classical motion planning problems, a robot is commanded to generate a trajectory to reach a target configuration without collision \cite{liu2022improved, dong2017experimental, douthwaite2019velocity}. Some works  extend the motion planning problems to  goals expressed in terms of temporal logic \cite{wongpiromsarn2013synthesis,lan2021survey, kantaros2020stylus}. They consider tasks with complex temporal and spatial constraints that can be expressed in a form closely linked to natural language. 
Linear temporal logic (LTL) is widely used to specify such tasks since it is a mathematically precise language with sufficient expressiveness \cite{clarke1997model}. Hybrid robot controllers \cite{lin2021hybrid} can be computed to satisfy tasks expressed in LTL formulae.

In classical controller synthesis approaches \cite{fainekos2009temporal, tian2021decentralized, schillinger2018simultaneous}, abstraction on the environment is created as a priori which simplifies the motion of robots into symbolic transitions. A  discrete plan can be  retrieved from a product automaton constructed by taking the product of the transition system and the task-related automata.
However, most existing works assume the environment is static, and its finite model can be obtained in advance. In the real world, the model of the environment is not always available and is even unpredictable due to shifting behavioral patterns. Directly resynthesis of the whole plan can lead to undesired consequences since the execution history will be lost, and the revised plan may violate the temporal specification. An example of a task is  ``Pick up a box in the warehouse first, then drop it off at the office before picking up another box.'' If the robot fails to locate the drop zone and decides to resynthesis the plan from scratch, the new plan will have the robot pick up a box again, which is prohibited by the task formula.

In case  the previous plan is unachievable, some  replanning methods are developed \cite{guo2013revising, livingston2012backtracking, lahijanian2016iterative} in recent years based on controller synthesis approaches.  
The work in \cite{guo2013revising} proposes a plan revising mechanism assuming the environment is partially known and static. Once one transition in the current plan is broken, it seeks ways to find the shortest path bridging up the two components again. 
Authors in \cite{livingston2012backtracking} develop a method to locally ``patch'' the defunct parts caused by the environmental change and identify new solutions. 
The work \cite{lahijanian2016iterative} develops an iterative repair strategy to resolve unexpected obstacles by combining local patching with refined triangulation. However, the above-mentioned repair operations cannot deal with both propositional change and moving obstacles, and replanning on large product automaton is 

Unlike the abstraction-dependent methods above,  our approach uses a sampling tree that explores  the workspace for solutions. The closest related work to this paper is \cite{luo2021abstraction}, which presents an abstraction-free method that incrementally builds trees to search for a satisfying solution. Note that it is an offline planner with no consideration of uncertainties.  A lot of researchers, either dealing with classical path planning problems \cite{gammell2014informed, strub2022adaptively} or temporal logic tasks \cite{luo2021abstraction},  put much effort into improving sample efficiency in sampling-based planning.  These methods invent various  heuristics to speed the convergence of the final solution to optimality. Some work \cite{karaman2011anytime, bircher2016receding} implement receding-horizon sampling trees  for interleaved planning and acting. However,  the executed edges are abandoned and cannot be used in future planning.   In \cite{naderi2015rt},  the authors develop a multi-query sampling-based path planner by preserving all nodes sampled.
Inspired by the approach above, we reuse the sampled points connected to the historical path for real-time replanning. 

We propose a novel sampling-based planning and replanning  paradigm for temporal logic tasks. 
The contributions of this work are threefold:
\begin{itemize}
	\item  A  temporal logic constrained path planner that rapidly responds to  real-time workspace knowledge updates using a  dual-root tree; 
	
	\item A  graph-theoretic approach that reuses sampled nodes and resolves state duplication during replanning; 

	\item Integration of algorithms with an online robotic perception-and-planning system and demonstration of its  advantages in simulation and real-world experiments. 
\end{itemize}

\section{Preliminaries}\label{sec prelimi}
\subsection{Linear Temporal Logic on Finite Traces}
 We restrict the task specification in the form of Linear Temporal Logic on finite traces ($\operatorname{LTL}_{f}$), which has sufficient expressiveness to command  robots with constrained power supply. 
Let AP be a set of atomic propositions where $\alpha \in A P$ is a Boolean variable. We consider $\operatorname{LTL}_{f}$ formulas whose syntax is as follows:
$$
\varphi:= \mathrm{true} \mid \alpha \mid \neg \varphi \mid  \varphi \vee \varphi \mid \varphi \wedge \varphi \mid \lozenge \varphi \mid  \square \varphi \mid  \varphi \mathcal{U} \varphi
$$
where   $\neg$ (negation), $\vee$ (disjunction), $\wedge$ (conjunction) are boolean operators, and $\lozenge$ (``eventually''), $\square$ (``always'') and $\mathcal{U}$ (``until'') are temporal operators. Note that we exclude the ``next'' operator  due to the continuous execution without instantaneous motion in our abstraction-free approach. We refer readers to the work \cite{de2013linear} for the semantics of $\operatorname{LTL}_{f}$.

For every $\operatorname{LTL}_{f}$ formula $\varphi$ there is a deterministic finite automaton (DFA) that accepts a sequence of input words to satisfy $\varphi$.
DFA is formally defined by a five-tuple $ \mathcal{A}_{\varphi}=\left(Q, 2^{A P}, \delta, Q_{0}, \mathcal{F}\right)$,
where $Q$ is a finite set of automata states; $Q_{0} \subseteq Q$ is a set of initial states; $2^{AP}$ is the input alphabet; $\delta: Q \times 2^{A P} \rightarrow 2^{Q}$ is the transition relation; $\mathcal{F} \subseteq Q$ is a set of accepting automata states.

 A  \textit{run} of a finite length $n$ on DFA is a sequence of states $r = q_{0}q_{1}\dots q_{n}$, where $q_{0} \in Q_{0}$ and $q_{k+1} \in \delta(q_k, \lambda) $ for some $\lambda \in 2^{AP}, k \in [0,n-1]$, which implies the run starts at an initial state and follows the transition relation. A run $r$ is called an \textit{accepting run} if $q_n \in \mathcal{F}$.

\subsection{Robot Model}
In this paper, we model a mobile robot as a hybrid system with state space $S = Q \times X$, where $Q$ is the set of discrete modes  and $X$ is the set of continuous states.  A \textit{hybrid state} is defined as a pair $s=(q,x) \in S$, and its projection on $X$ is denoted as $s|_{X} = x$. Considering no external disturbance caused by the environment, the evolution of a robot can be achieved by applying a motion primitive $u$ during  time $t$ at a state in the domain $x \in X$, which can be written as $\zeta(x, u, t)$.   A finite \textit{path} is a finite sequence of hybrid states $\tau = s_{0}s_{1}\dots s_{n}$, and its projection $\tau|_{X}$ is resulted from a series of motion primitives  
$\zeta(x_{0}, u_{0}, t_{0})\zeta(x_{1}, u_{1}, t_{1}) \dots\zeta(x_{n}, u_{n}, t_{n})$. 
The corresponding \textit{trace} of a path $\tau$ is $\sigma = L(x_{0})L(x_{1})\dots L(x_n)$, where $L:X \rightarrow 2^{AP}$ is a labeling function that returns the atomic propositions that are true in the state $x \in X$. 
We define that a path $\tau$ satisfies an $\operatorname{LTL}_{f}$ formula $\varphi$ if and only if the input words of the corresponding trace lead to an acceptance state, formally written as $\exists \ q_{f} =  \delta\left(q_{n}, L(x_{n}) \right), q_{f} \in \mathcal{F} \Leftrightarrow \tau \models \varphi$. 
 
Follow  \cite{lahijanian2016iterative},  an \textit{event-driven trace} of a path is defined as 
$$
\bar{\sigma}=L\left(x_{0}\right) L\left(x_{i_{0}}\right) L\left(x_{i_{0}+i_{1}}\right) \ldots L\left(x_{i_{0}+\cdots+i_{l-1}}\right),
$$
for $i_{0}+\cdots+i_{l-1}<n$ by removing states with repetitive discrete modes in a path. In another word, the \textit{event-driven trace} captures the label change in the trajectory and $L(x_{I_j}) \neq L(x_{I_{j+1}})$ where $I_{j}=\sum_{k=0}^{j} i_{k}$.  In practice, there may exist multiple paths leading to the acceptance of a DFA with different  \textit{event-driven traces}.

\section{Problem Description and Proposed Approach} \label{sec framework}

\subsection{Problem Description}
This paper considers a mobile robot with sensory capability moving in a partially unknown workspace to accomplish high-level goals. The mobile robot can either be a differential drive ground vehicle or a multicopter, which can remain stationary and change its heading. {Meanwhile, the robot is not collision resilient and is forced to stay away from obstacles by the planner. Therefore, collision avoidance constraints are excluded from formulae in the following sections.}  The workspace $X$ consists of a set of regions of interest $\Pi$ whose properties are the combinations of atomic propositions $AP$, and a set of regions occupied by fixed obstacles $X_{\mathrm{occ}}$.
The robot is given the workspace model $X_{\mathrm{init}}$  as prior knowledge.  We assume $X_{\mathrm{init}}$ suffices the robot to compute a trajectory within obstacle-free areas $X_{\mathrm{free}} = X_{\mathrm{init}} \backslash X_{\mathrm{occ}}$ that satisfies the specification $\varphi$. In contrast to the work \cite{lahijanian2016iterative},   we relaxed the assumption that the propositional regions are fully-known.  Due to the incomplete information given in advance, satisfaction is not guaranteed since the actual region property might not coincide with that in  prior knowledge.  The workspace could be updated by receiving knowledge from either onboard sensors or external observers.

 The problem that this paper focused on is stated as follows:
 
\begin{Problem} \label{prob main}
Given a partially unknown workspace $X_{\mathrm{init}}$ with dynamic obstacles, a  hybrid robot model whose state space is $S$, and a task specification $\varphi$ expressed in $\operatorname{LTL}_{f}$, 
find a path $\tau$ that satisfies  $\varphi$ with adaptation to any admissible uncertainties in real-time.
\end{Problem}

 We suppose that the uncertainty in the environment is  admissible in the sense that the propositions of each region do not vary dynamically, and there always exists a collision-free finite  path that results in the task's acceptance. Once detected property changes or dynamic obstacles,  the robot is supposed to be able to brake and avoid reaching/colliding with them in time while satisfying  kinodynamic constraints. We assume no perception error within a sensing area, while  actuation uncertainties are taken care of by  feedback motion planning at a lower level.

\subsection{Proposed Approach}
\begin{figure}[htb]
	\centering
	\setlength{\belowcaptionskip}{-0.3cm}
    \setlength{\abovecaptionskip}{0.1cm}
	\includegraphics[width=0.5\textwidth]{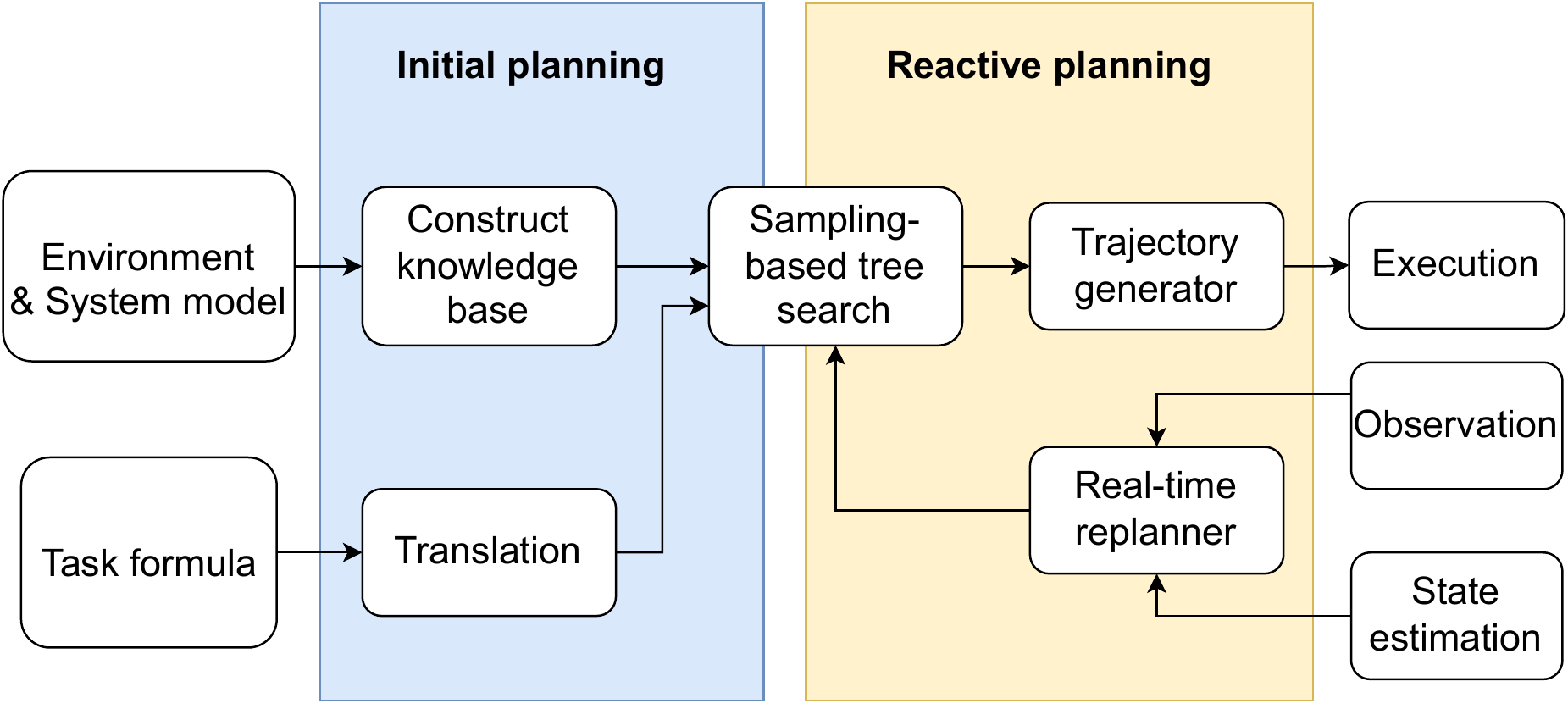}
	\caption{The framework of real-time temporal logic planning}
	\label{fig:sys}
\end{figure}

To address Problem~\ref{prob main}, we propose a path planning framework whose diagram can be viewed in Fig.~\ref{fig:sys}. The planning part can be divided into two phases: initial planning and reactive planning. 

Prior to planning, the system is provided with an environment model, a robot model, and a task formula. The robot model helps grow a  sampling-based search tree $\mathcal{T}$ in $X_\mathrm{free}$ progressively with kinetically compatible edges. 
 The task  specified in $\operatorname{LTL}_{f}$ as $\varphi$ is translated into a DFA and is further converted to a state monitor that guides the sampling-tree expansion. A knowledge base of the workspace is constructed based on the environment model with labeled regions $\Pi$ and obstacles $X_{\mathrm{occ}}$. The knowledge base serves as an initial map $X_{\mathrm{init}}$ in the initial planning to obtain an initial solution whose trace satisfies the task formula $\varphi$,{ as to be explained in Section~\ref{sec initplan}.}

In the online reactive planning part, once a solution is successfully obtained, the robot can track the nominal path safely with the help of a  trajectory generator. During   execution, the sampling-based planner continuously searches for a distinctive solution and improves the existing solutions,  {as will be described in Section~\ref{sec sol_simprove}}. The state estimation helps the planner keep track of the execution progress and updates the robot's continuous state $x$. In the meantime, the robot receives observations from both external observers and its onboard sensors and revises the corresponding part of the knowledge base.   The evolution of both the environment state and the robot state can trigger the replanning mechanism to deal with the non-determinism in  the domain, {as will be discussed in Section~\ref{sec rbt_state} and Section~\ref{chap percept}. }

\section{ Solution of the Problem} \label{sec alg}
\subsection{Automata preprocess and temporally forbidden zones} \label{sec preprocess}
 To obtain a minimized automaton without ambiguity, we use MONA \cite{ijcai2017-189} and $\operatorname{LTL}_{f}$2DFA \cite{de2018automata} to translate an $\operatorname{LTL}_{f}$ formula to a DFA $ \mathcal{A}_{\varphi}$. 
Chances are that, given a prefix of a run $ q_{0}q_{1}\dots q_{i}$ on DFA, there does not exist a sequence $ q_{i}q_{i+1}\dots q_{n}$ where $q_{k+1} \in \delta(q_k, \lambda) $ for any $\lambda \in 2^{AP}, k \in [i,n-1]$ and $q_n \in \mathcal{F}$, i.e., the accepting states cannot be reached from the automata state $q_{i}$ on the DFA.    To overcome this issue, we apply a preprocessing routine on DFA before planning. In detail, we construct an auxiliary automaton that reverses all transitions of the original DFA and finds the sets of automata states $\mathcal{B}$ that are not reachable from any state in $\mathcal{F}$. During planning, the planner disables any transition relation $ \delta(q_k, \lambda)$ that can have the motion tree to include a node with an automata state $q_{k+1} \in  \mathcal{B}$.

During execution, to ensure the states in $\mathcal{B}$ are never reached, temporally forbidden zones $\Pi_{\mathrm{fbd}}$ are extracted and serve as time-variant hard constraints in path tracking. When the robot's state is $s_{\mathrm{cur}} = (q_{\mathrm{cur}}, x_{\mathrm{cur}})$, the planner obtains the next discrete mode  as $q_{\mathrm{nxt}} = \delta(q_{\mathrm{cur}}, L(x_{\mathrm{cur}})) $ due to the determinism of DFA. 
The planner examines each labeled region $\pi \in \Pi$, and tries to intake the corresponding label $L(\pi) $. If the input word transitions $q_{\mathrm{nxt}}$ toward a state in $\mathcal{B}$,  the corresponding region will be added into $\Pi_{\mathrm{fbd}}$.

\subsection{Initial plan computation} \label{sec initplan}
After constructing the initial map $X_{\mathrm{init}}$ and translating the $\operatorname{LTL}_{f}$ specification, the robot builds a motion tree toward the temporal goal. Similar to anytime RRT* \cite{karaman2011anytime}, the robot does not wait till a near-optimal solution is calculated to execute. Instead, once a feasible solution is found, the robot begins to follow the nominal path with a trajectory generator. 

Like the work \cite{luo2021abstraction}, we employ a sampling-based tree search method monitored by an automaton. The search tree $\mathcal{T}$ is composed of node set $\mathcal{V}_{\mathcal{T}}$ and edge set $\mathcal{E}_{\mathcal{T}} $. Additionally, we maintain a graph $\mathcal{G} = (\mathcal{T}, \mathcal{V}_{\mathrm{iso}})$  containing not only the search tree but also a set of isolated vertices.

Every time a sample is drawn on a state $x_{\mathrm{rand}} \in X_{\mathrm{free}}$, the nearest node in the tree $s_{\mathrm{nearest}} = (q_{\mathrm{nearest}}, x_{\mathrm{nearest}}) $ steers to $x_{\mathrm{rand}}$ with robot dynamics in time $\Delta t$ and ends at $x_{\mathrm{new}}$.  By pairing with possible automata states,  a set of hybrid states $S_{\mathrm{new}} = \{(q, x_{\mathrm{new}}) \mid q \in Q \backslash \mathcal{B} \}$ whose continuous states are the same but automata states are different  is generated. 
 $S_{\mathrm{new}}$ is recorded in each new vertex $s_{\mathrm{new}} \in S_{\mathrm{new}}$ for resolving state duplication and will be referred to as  $S^{\mathrm{rcd}}$ in Section~\ref{sec rbt_state}. 
Afterward, the planner tries to add the hybrid states as nodes to the tree. The set of vertices  that is near $x_{\mathrm{new}}$ is $S_{\mathrm{near}} = \{s=(q, x) \in  \mathcal{V}_{\mathcal{T}} \cup \mathcal{V}_{\mathrm{iso}} \mid || x-x_{\mathrm{new}} || \leq r_{n} (\mathcal{V}_{\mathcal{T}}) \}$, where $r_{n}$ is a radius calculation function as defined in \cite{luo2021abstraction}. For each $s_{\mathrm{new}}=(q_{\mathrm{new}}, x_{\mathrm{new}})$, 
we find the node $s_{\mathrm{best}} = (q_{\mathrm{best}}, x_{\mathrm{best}}) \in S_{\mathrm{near}}$ with the least cost-to-come value that satisfies the transition relation $q_{\mathrm{new}} = \delta\left(q_{\mathrm{best}}, L(x_{\mathrm{best}}) \right)$, and then add an edge $\varepsilon = <s_{\mathrm{best}}, s_{\mathrm{new}}>$ to the tree $\mathcal{T}$. 
The \textit{cost-to-come values}—the cumulative distance from the tree root to the states—are allocated to the newly added nodes $S_{\mathrm{added}}$ in the tree.  The remaining nodes   $ S_{\mathrm{new}} \backslash S_{\mathrm{added}}$ are included in graph $\mathcal{G}$, and they can be added to the tree in the future by rewiring. The rewiring process is similar to that in \cite{naderi2015rt}
except for the hybrid state space and  constraints on transition relations. Note that not only tree nodes $\mathcal{V}_{\mathcal{T}}$ but also isolated vertices $V_{\mathrm{iso}}$ can be rewired. The planner checks if the newly sampled node $s_{\mathrm{new}}=(q_{\mathrm{new}}, x_{\mathrm{new}})$ can lead to the acceptance one hop away: If $\delta (q_{\mathrm{new}}, L(x_{\mathrm{new}})) \in \mathcal{F} $, then a new solution is identified immediately.  

\subsection{Real-time solution improvement} \label{sec sol_simprove}

Since the workspace may be very large, the initial feasible plan can be sub-optimal and has a large accumulative cost value. We develop a solution update scheme that continuously improves the solution and steers the robot following the least-cost path toward the temporal goal. 

Every time an accepting path $\tau$ is found,  the planner sets $\tau$ as the current solution if no solutions have been found so far. Otherwise, the planner inspects the solutions obtained before and only updates those  solutions that are similar to the new solution while costing more. The intuition of the procedure above is that different from  traditional path planning whose goal is a point in  cartesian space, a temporal logic goal can be achieved as long as the corresponding run is accepting. 
An example is ``go to site A or site B,'' where distinct paths driving the robot to either site A or site B can both fulfill the given task. To exclude duplicated solutions, the planner compares two solutions on their \textit{event-based traces} as well as  path topology. We use the concept of  homotopy class \cite{bhattacharya2012topological} to decide whether paths are similar in topology, i.e., whether they are with the same start and end positions and can be deformed into each other without crossing obstacles or labeled regions.
The set of paths that satisfies the temporal goal and are distinct from each other are stored in a solution library. The best solution is elected from all path candidates with the least cost.

\subsection{Robot state update} \label{sec rbt_state}
\begin{figure}[htb]
	\centering
	\setlength{\belowcaptionskip}{-0.2cm}
    \setlength{\abovecaptionskip}{0.1cm}
	\includegraphics[width=0.5\textwidth]{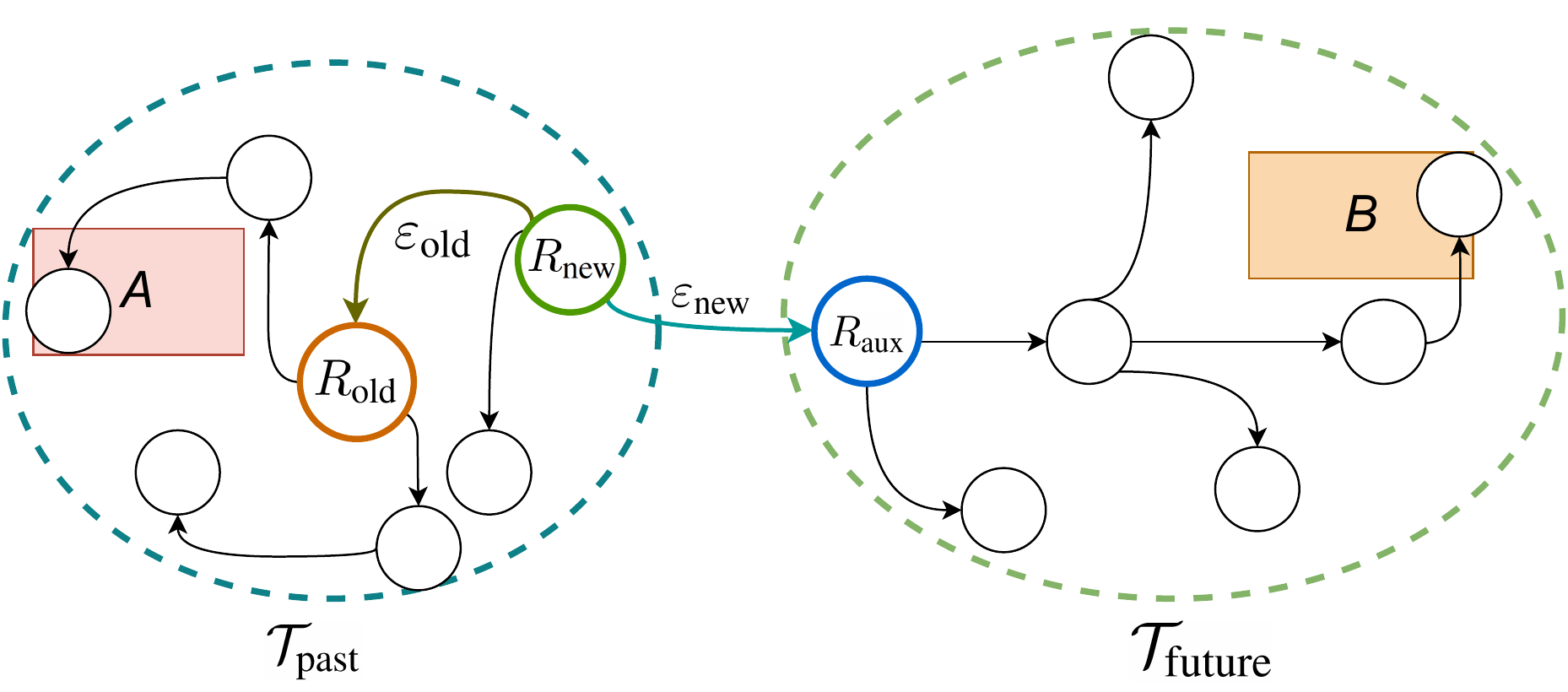}
	\caption{Illustration of a dual-root tree during execution. The motion tree $\mathcal{T}$ is composed by $\mathcal{T}_{\mathrm{past}} \cup \varepsilon_{\mathrm{new}} \cup \mathcal{T}_{\mathrm{future}}$. Both $R_{\mathrm{new}}$ and $R_{\mathrm{aux}}$ have zero cost-to-come values.}
	\label{fig:dualroot}
\end{figure}

In existing works on anytime path planning \cite{karaman2011anytime, westbrook2020anytime}, the planning only focuses on the pending portion of the current plan while the nodes traveled are not useful anymore. However, considering the potential incidents in the real world, the robot may have to turn around and travel through the visited nodes in order to satisfy  temporal logic tasks. During the execution of the plan, we keep the traveled edges alive by moving the root along with the robot motion. 
Instead of setting the next un-visited node as root like in \cite{karaman2011anytime,naderi2015rt},  we transform the classical tree into a dual-root tree during the robot movement.
Fig.~\ref{fig:dualroot} uses the task ``go to site A or site B'' as an illustration.
Denoting the executed edge connecting the last root $R_{\mathrm{old}}$ and the current motion target $s_{\mathrm{tgtOld}}$ as $\varepsilon_{\mathrm{old}} = <R_{\mathrm{old}}, s_{\mathrm{tgtOld}} >$, we reverse this edge once the robot arrives at $s_{\mathrm{tgtOld}}$. Then $s_{\mathrm{tgtOld}}$ is set as the new root of the tree $R_{\mathrm{new}}$. We further denote the current edge to execute  as $\varepsilon_{\mathrm{new}} = <R_{\mathrm{new}}, s_{\mathrm{tgtNew}} >$, the subtree stemming from $s_{\mathrm{tgtNew}}$  as $\mathcal{T}_{\mathrm{future}}$, and the part $\mathcal{T} \backslash  \mathcal{T}_{\mathrm{future}} \backslash \varepsilon_{\mathrm{new}}$ as $\mathcal{T}_{\mathrm{past}}$.  $\mathcal{T}_{\mathrm{past}}$ will not be visited  as the execution progresses on the current solution unless the replanning is triggered. A rewire process starts from $R_{\mathrm{new}}$ at every timestamp to minimize the cost-to-come values of each node in $\mathcal{T}_{\mathrm{past}}$. Additionally, $s_{\mathrm{tgtNew}}$ is treated as an auxiliary root $R_{\mathrm{aux}}$, and the cost-to-come value of every node in $\mathcal{T}_{\mathrm{future}}$ is calculated by retracing to $R_{\mathrm{aux}}$. This is an essential step to prevent nodes in  $\mathcal{T}_{\mathrm{future}}$ be rewired to $\mathcal{T}_{\mathrm{past}}$ during execution. 

\vspace{-0.3cm}

One can notice that when the edge $\varepsilon_{\mathrm{old}} $ is reverted, the transition relation $R_{\mathrm{old}}|_{Q} = \delta\left(R_{\mathrm{new}}|_{Q}, R_{\mathrm{new}}|_{X} \right)$ might not hold. Hence Algorithm~\ref{alg propAstate} is applied on $\mathcal{T}_{\mathrm{past}}$ in a depth-first-search manner. 
At the beginning of the propagation,  $R_{\mathrm{new}}$ is the parent node, and $R_{\mathrm{old}}$ is the child node.  
In Line \ref{line checkatran}, the function CheckATran($\cdot$) checks if the transition relation from $q_m$ to $q_n$ is valid given the label of the parent node. If the transition is broken, routines described in Line \ref{line succ()}-\ref{line propgated} will be applied. The algorithm searches the set of the successor states of $ q_{m}$ and finds the state $ q_{k}$, which can be evolved from $ q_{m}$ by taking the word $ L(x_{i})$. 
The automata state of the corresponding node is therefore updated.   This process ensures that all nodes in $\mathcal{T}_{\mathrm{past}}$ are reachable in future replanning. Recall the sampling process in Section~\ref{sec initplan}, $s_{c}$ records a node set $S^\mathrm{rcd}_{c}$ with all possible combinations of $x_{c}$ and $q \in Q \backslash \mathcal{B}$. If there is no node in $S^\mathrm{rcd}_{c}$ with the same hybrid state as $s_{c}$ (Line~\ref{line samestate}), the algorithm adds a copy of $s_{c}$ to the isolated vertices set $\mathcal{V}_{\mathrm{iso}}$ (Line~\ref{line nodecpy}) before modifying $s_{c}$. 

\begin{algorithm}\footnotesize
\caption{PropagateState}\label{alg propAstate}
\LinesNumbered 
\KwIn{Parent node $s_{p} $, child node $ s_{c}$, DFA $ \mathcal{A}_{\varphi}$, isolated vertices set $\mathcal{V}_{\mathrm{iso}}$}
\KwOut{Successfulness of the propagation }

  $s_{p} = (q_{m}, x_i) $\\
  $ s_{c}= (q_{n}, x_j) $\\
  \If{$\mathsf{CheckATran}(q_{m}, L(x_{i}), q_{n}, \mathcal{A}_{\varphi})$}{ \label{line checkatran}
  \Return $\top$
  
  }
\For{$q_{k} \in \left\{q_{v} \mid \exists \lambda \in 2^{A P}, q_{v} = \delta\left(q_{m}, \lambda \right)\right\}$}{ \label{line succ()}
    \If{$\mathsf{CheckATran}(q_{m}, L(x_{i}), q_{k}, \mathcal{A}_{\varphi})$}{  
        \If{$|\mathsf{SameStateNodes}(s_{c})| = 0 $ }{ \label{line samestate}
            $\mathsf{AddNodeCopy}(s_{c}, \mathcal{V}_{\mathrm{iso}}) $\\ \label{line nodecpy}
        }
        $ s_{c} \gets (q_{k}, x_j) $\\ \label{line modifystate}
         \Return $\top$ \label{line propgated}
        }
    }

\Return $\perp$

\end{algorithm}

However, after line~\ref{line modifystate} in Algorithm~\ref{alg propAstate} modifies the automata state of $s_{c}$,  duplication may occur between  $s_{c}$ and the node with the same hybrid states in $S^\mathrm{rcd}_{c}$. Similar problems occur after the propagation in dealing with propositional updates, 
as to be mentioned in Section~\ref{chap percept}. 
It will incur problems in extending and rewiring nodes as several nodes with identical hybrid states and costs exist simultaneously. Besides, duplicated nodes  make it difficult to maintain tree sparsity. To prevent the tree from becoming entangled and excessively dense, we require an algorithm to visit all the duplicated nodes and only keep the one that has the least cost-to-come value.

 \begin{algorithm}\footnotesize
\caption{MergeNode}\label{alg mergeNode}
\LinesNumbered 
\KwIn{Node $s_{m} $, tree vertices set $\mathcal{V}_{\mathcal{T}}$, tree edge set $\mathcal{E}_{\mathcal{T}}$,  isolated vertices set $\mathcal{V}_{\mathrm{iso}}$}
\KwOut{Successfulness of the merging }

 $S_{\mathrm{dup}} \gets \mathsf{SameStateNodes}(s_{m})$\\ \label{line:getsame}
 $\mathcal{V}_{\mathrm{isoDup}} \gets S_{\mathrm{dup}} \cap \mathcal{V}_{\mathrm{iso}}$ \\ \label{line:removeiso1}
 $\mathcal{V}_{\mathrm{iso}} \gets \mathcal{V}_{\mathrm{iso}}  \backslash \{ \mathcal{V}_{\mathrm{isoDup}} \}$ \\
$S_{\mathrm{dup}} \gets S_{\mathrm{dup}} \backslash \{ \mathcal{V}_{\mathrm{isoDup}} \}$\\ \label{line:removeiso2}
 \If{$|S_{\mathrm{dup}} | =0$ }{
 \Return $\top$
 }
  $s_{d} \gets S_{\mathrm{dup}}.\text{pop} () $ \\ \label{line get first}
  \If{$s_{d} = \mathsf{Parent}(s_{m}) \vee \mathsf{Cost}(s_{d}) \leq \mathsf{Cost}(s_{m})$}{
  $s_{\mathrm{better}} \gets s_{d} $\\
  $s_{\mathrm{worse}} \gets s_{m} $
  }\Else{  
  $s_{\mathrm{better}} \gets s_{m} $\\
  $s_{\mathrm{worse}} \gets s_{d} $}
  
  $s_{\mathrm{better}}.\textrm{visited}  \gets \top$ \\
  $s_{\mathrm{worse}}.\textrm{visited} \gets \top$ \\
\For{$s_{\mathrm{child}} \in \mathsf{Children} (s_{\mathrm{worse}})$}{\label{line:reconnect1}
    $\mathcal{E}_{\mathcal{T}} \gets \mathcal{E}_{\mathcal{T}} \backslash \{ < s_{\mathrm{worse}}, s_{\mathrm{child}}>  \} $ \\
    $\mathcal{E}_{\mathcal{T}} \gets \mathcal{E}_{\mathcal{T}} \cup \{ < s_{\mathrm{better}}, s_{\mathrm{child}}>  \} $ \\
    }
        $\mathcal{E}_{\mathcal{T}} \gets \mathcal{E}_{\mathcal{T}} \backslash \{ < \mathsf{Parent}(s_{\mathrm{worse}}), s_{\mathrm{worse}}>  \} $ \\  
        $\mathcal{V}_{\mathcal{T}}  \gets \mathcal{V}_{\mathcal{T}} \backslash \{ s_{\mathrm{worse}} \}$\\ \label{line:reconnect2}
    
\Return $\top$

\end{algorithm}

We propose Algorithm~\ref{alg mergeNode} to deduplicate the nodes in linear time. When handling robot state updates, the algorithm is recursively applied on the $\mathcal{T}_{\mathrm{past}}$ in a preorder fashion and eliminates one duplicated node at a time.   In Line \ref{line:getsame}, denoting the visiting node as $s_{m}$,  all the nodes with the same hybrid state  are identified as $S_{\mathrm{dup}}$ given the  node set  $S^{\mathrm{rcd}}_m = \{ s \in  \mathcal{G} \mid s|_{X} = s_{m}|_{X} \}$. In Line~\ref{line:removeiso1}-Line~\ref{line:removeiso2},  the duplicated and isolated states  $\mathcal{V}_{\mathrm{isoDup}} $ are detected and be removed from the isolated vertices set $\mathcal{V}_{\mathrm{iso}}$ as well as duplicated states set $S_{\mathrm{dup}}$ permanently.   In Line~\ref{line get first}, one  element $s_{d}$ in the set $S_{\mathrm{dup}}$ is selected, while the rest of the nodes are to be processed later. If the pair of states $s_{d},s_m $ is directly connected, i.e., are in parent-child relationship, the child node is  marked as $s_{\mathrm{worse}}$. Otherwise, the algorithm evaluates the cost of the two nodes $ s_{m}, s_{d} $ and marks the node that costs more as $s_{\mathrm{worse}}$. After that, in Line~\ref{line:reconnect1}-Line~\ref{line:reconnect2}, the inward  and outward edges of $s_{\mathrm{worse}}$ is re-connected to the better node $s_{\mathrm{better}}$, and $s_{\mathrm{worse}}$ node is safely deleted. The offspring of the $s_{\mathrm{better}}$ is visited recursively by  Algorithm \ref{alg mergeNode} until every node in the subtree is marked as visited.

As a consequence of Algorithm~\ref{alg propAstate} and Algorithm~\ref{alg mergeNode} together, all nodes in  $\mathcal{T}$ are updated with future-reachable automata states, and the cost-to-come value of each node is  less than or equal to the original value.

\subsection{Environment state update} \label{chap percept}
In real-world applications, prior knowledge may not be precise, making the acceptance condition $\tau \models \varphi$ invalid. To tackle this problem, we enable the robot to sense the environment and receive observations from other agents. We consider a mobile robot moving in a scenario where the properties of regions can change, and unexpected obstacles may block the path. The robot continuously  analyzes sensor data and extracts \textit{knowledge} in two forms: $\kappa_{\mathrm{reg}} = (\pi,\ L(\pi)) $ or $\kappa_{\mathrm{obs}} = (\upsilon, \ x_{\upsilon}) $,
where $\pi \in \Pi$  stands for the region covered by the sensor, $L(\pi)$  is the set of propositions satisfied in $\forall x \in \pi$,   $\upsilon$ is an unexpected obstacle instance, and $x_{\upsilon}$ is the position of $\upsilon$. If knowledge stored in memory is found inconsistent with the newly obtained knowledge,  the robot will update the knowledge base and perform plan revision.

Every time $\kappa_{\mathrm{reg}}$ is obtained, labels of the region $\pi$ are updated. The planner first  calls Algorithm~\ref{alg propAstate} to correct the states of nodes in the whole tree $\mathcal{T}$ given the updated workspace, and then uses  Algorithm~\ref{alg mergeNode} to resolve state duplication. Afterward, each possible solution  is examined to see if  its trace satisfies $\varphi$. If there still exist feasible solutions, the solution with the least accumulative cost is set as the current plan. Otherwise, if the knowledge change is critical and all previous solutions are infeasible, the robot will come to a full stop at $x_{\mathrm{stop}}$. The new root is set as $s_{\mathrm{stop}} = (q_{\mathrm{stop}}, x_{\mathrm{stop}}), $ where $q_{\mathrm{stop}} \in Q$  is calculated by the previous root. The robot waits at $x_{\mathrm{stop}}$ until the planner obtains a new solution as described in Section~\ref{sec initplan}.

Given  knowledge $\kappa_{\mathrm{obs}}$, if the obstacle $\upsilon$ is not observed before,  or the traveled distance $\Delta x_{\upsilon}$  exceeds a threshold, the planner updates $X_{\mathrm{free}}$ and triggers the replanning procedure.  The planner first blocks the edges  that cross $\upsilon$  by setting their weights of them as positive infinity. 
Afterward, like the situation in dealing with $\kappa_{\mathrm{reg}}$, the robot will carry out the best executable solution or await a feasible solution generated by the sampling-based search. 

  After  coping with  environmental state change, the search tree is corrected so that every edge in $\mathcal{T}$ satisfies its transition relation in $\mathcal{A}_\varphi$. If no solutions have been found so far, the planning is converted to an LTL-constrained sampling-based tree search  described in the work \cite{luo2021abstraction}, which has been proved to be  probabilistically complete. Hence by continually growing $\mathcal{T}$ during replanning, a solution that satisfies $\varphi$ is bound to be found  if  one exists in the domain $X$. 
Therefore, we can conclude that Problem~\ref{prob main} is solved.

\section{Implementation and Experimental Results} \label{chap res}

For comparison and illustration purposes, we present case studies on fire extinguishing tasks.
The target scenario contains three regions of interest, three known obstacles, and one unknown obstacle, {as can be seen in  Fig.~\subref*{fig:kreg} and Fig.~\subref*{fig:kobs}. Specifically,  $l_{1}$ and $l_{3}$ are labeled as grasslands and $l_{2}$ is a pond in the robot's prior knowledge, and the robot can validate a label or detect an obstacle within its sensing area (3m$\times$3m).  
To ensure that the sensing distance is enough for braking, we set the maximum velocity of the robot and the obstacle as 0.5~m/s and 0.2~m/s, respectively.}

\subsection{Comparative results in simulation}
We conduct simulations in three workspaces with different types of uncertainties. $X_A$ is a workspace where the initial knowledge of either one of \textit{grassland} regions is inaccurate. $X_B$ is a workspace where one obstacle moves from one side of the workspace to the other side. $X_C$ combines the property uncertainty and occupancy uncertainty of $X_A$ and $X_B$.
A vehicle is given a task formula
$\varphi_{1}=\lozenge (pond \wedge \lozenge grassland)
,$
which instructs the vehicle to fetch water from the pond and then put out the fire in the grasslands.  
We select  methods from   Guo et al. \cite{guo2013revising} and Luo et al.  \cite{luo2021abstraction}  to compare the replanning speed and solution quality, and all simulations are conducted in Python with an Intel i7  CPU.  The abstraction-dependent approach \cite{guo2013revising} is used to compare the performance only in $X_A$ since the planner cannot deal with dynamic behaviors in $X_B$ and $X_C$. The abstraction-free approach \cite{luo2021abstraction} is implemented by us in a reactive way that, once the current path is found invalid, the vehicle saves the execution progress and builds the search tree again.   Results  can be viewed in Table~\ref{tab:diff-prop}, where each entry is obtained by averaging 30 simulation rounds with randomized initial robot configurations.  
Our planner outperformed Luo et al. \cite{luo2021abstraction} by adapting to uncertainties faster and planning shorter paths to satisfy temporal goals. Furthermore, our approach has broader applicability than Guo et al. \cite{guo2013revising}.
\begin{table}[htbp]
		\centering
		\footnotesize
		\caption{Performance of different planners }
		\label{tab:diff-prop}
		\tabcolsep 3pt 
		\renewcommand\arraystretch{1.2}

\begin{tabular}{|c|c|c|c|c|}
\hline
Metrics                                      & Methods   & $X_A$  & $X_B$ & $X_C$ \\ \hline
\multirow{3}{*}{Total Completion Time (s)} &  Guo et al.         &  23.15                  &       ---          &   ---     \\ \cline{2-5} 
& Luo et al. &        30.19       &     34.78            &  46.94     \\ \cline{2-5} 
                                             & Ours      &         \textbf{ 22.90 }        &     \textbf{30.14 }           &   \textbf{29.56 }   \\ \hline
\multirow{3}{*}{Average Replanning Time (s)} &   Guo et al.        &    0.44                &  ---               &     ---   \\ \cline{2-5}                   
& Luo et al. &        5.87         &       6.82          &   6.92    \\ \cline{2-5} 
                                             & Ours      &       \textbf{  0.39  }        &     \textbf{ 1.63 }          &  \textbf{1.24 }   \\ \hline
\multirow{3}{*}{Total Travel Distance (m)} &   Guo et al.        &       10.0           &      ---            &   ---     \\ \cline{2-5} 
& Luo et al. &       8.49          &       8.54          & 9.83      \\ \cline{2-5} 
                                             & Ours      &        \textbf{ 8.01  }         &    \textbf{ 7.03 }           &  \textbf{9.26  }   \\ \hline

\end{tabular}
\end{table}

\subsection{System integration and real-world experiment} 
The  proposed planner produces a  path $\tau = s_{0}s_{1} \dots s_{n} $ satisfying the task formula. To drive the robot ahead following the piece-wise linear path $\tau | _{X}$, a trajectory generator is required to track the path without deviating from the task formula. Note the trajectory generator is system-dependent due to different robot dynamics. Following the assumption of robot dynamics in Section~\ref{sec framework}, we utilize a motion-primitive-based local motion planner \cite{lai2019model}, which can be deployed on both multicopter and differential-drive vehicles. In detail, given start configuration $x_{i}$ and end configuration $x_{i+1}$, the motion planner solves a boundary value problem from $x_{i}$ to $x_{i+1}$ and generates a motion primitive  $ \zeta(x_{i}, u_{i}, t_{i})$ { in a model predictive control manner}. In the meantime, a volumetric mapper \cite{chen2022gpu} builds a Euclidean distance transform (EDT) map for the motion planner by combining data from prior knowledge and  online observation.

\begin{figure}[htb]
\setlength{\belowcaptionskip}{-0.2cm}
\setlength{\abovecaptionskip}{0.1cm}
	\centering	\includegraphics[width=0.35\textwidth]{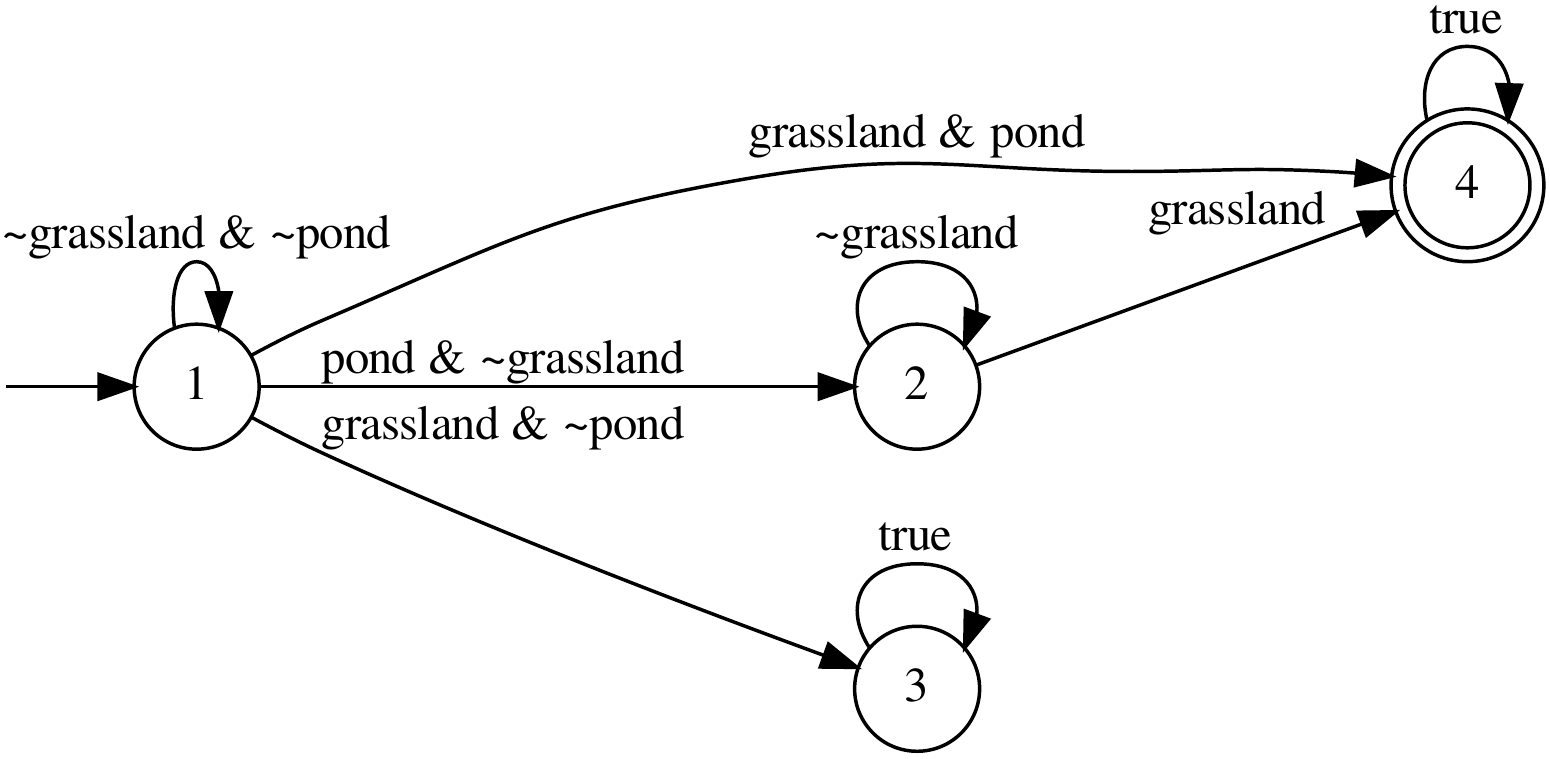}
	\caption{The DFA of  task formula $\varphi_{2}$}
	\label{fig:aut}
\end{figure}

In the real-world experiment, we use a multicopter to follow the nominal path in real time. A motion capture system provides  position estimation for the multicopter as well as the observation of  the obstacle if  within the sensing range. 
The multicopter is asked to accomplish the task: `` Don't fly over the grasslands until you've got water from the pond, and eventually put out the fire in a grassland without colliding with obstacles.'' The formula is written as
$\varphi_{2}=(\neg grassland \ \mathcal{U} \ pond)\wedge \lozenge grassland , $ and is translated into a DFA, as in Fig.~\ref{fig:aut}.

\vspace{-0.3cm}
In DFA preprocess (Section~\ref{sec preprocess}), the algorithm identifies that the automata state \textbf{3}  belongs to  $\mathcal{B}$.
 The resulting temporally forbidden zones $\Pi_{\mathrm{fbd}}$ together with unexpected obstacles  are encapsulated into pointcloud and broadcast to ROS. The volumetric mapper clusters the pointcloud and yields an EDT map pushing the robot away from $\Pi_{\mathrm{fbd}}$ and obstacles.
With the EDT map updated in every planning iteration, the motion planner generates a series of safety-guaranteed motion primitives.
All computations are running at 10 Hz on an onboard computer, Nvidia Xavier NX.
\begin{figure}[htb]
\setlength{\belowcaptionskip}{-0.3cm}
\setlength{\abovecaptionskip}{0.1cm}
		\subfloat[]
{\includegraphics[width=0.22\textwidth]{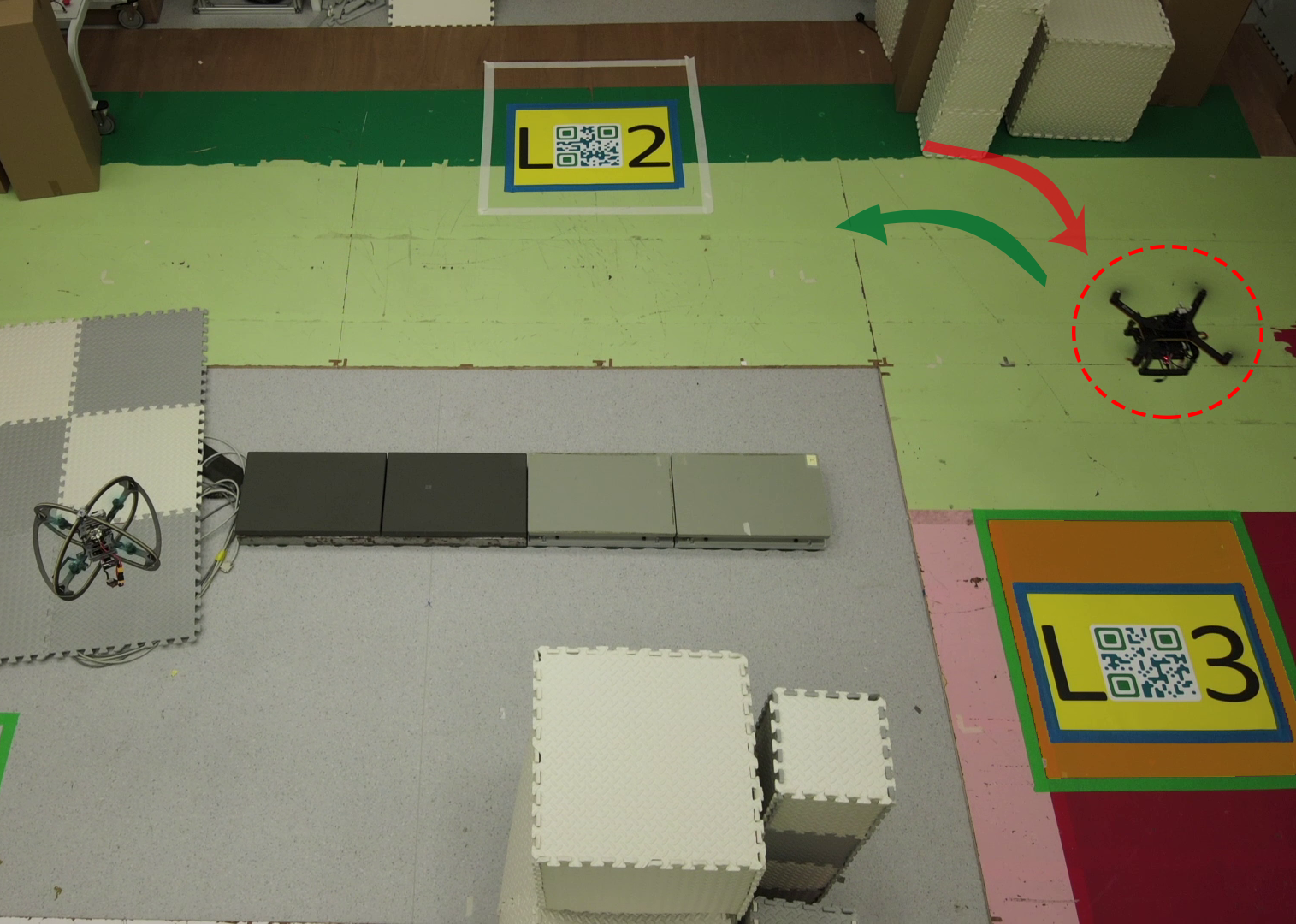} \label{fig:kreg}
}\hfill
		\subfloat[]
{\includegraphics[width=0.22\textwidth]{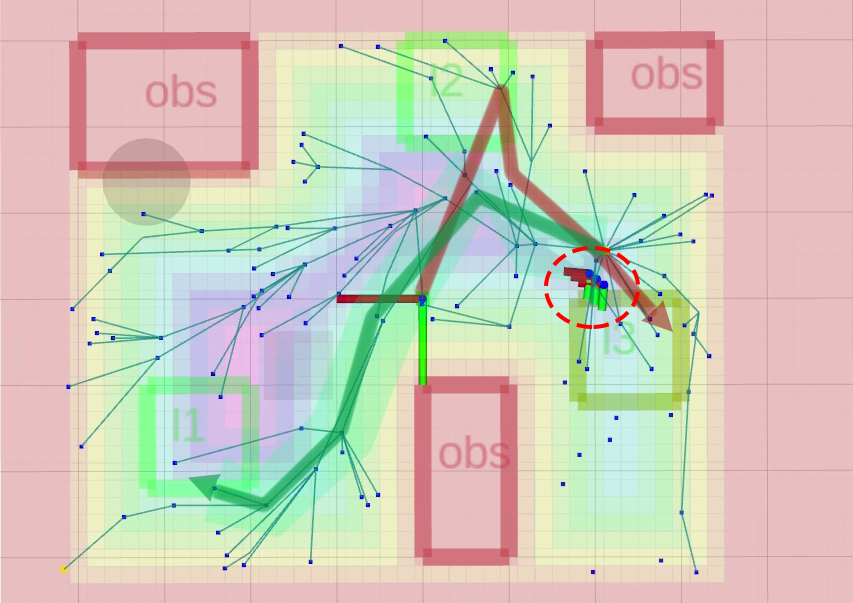} \label{fig:replanRegJT}
}\hfill
		\subfloat[]
{\includegraphics[width=0.22\textwidth]{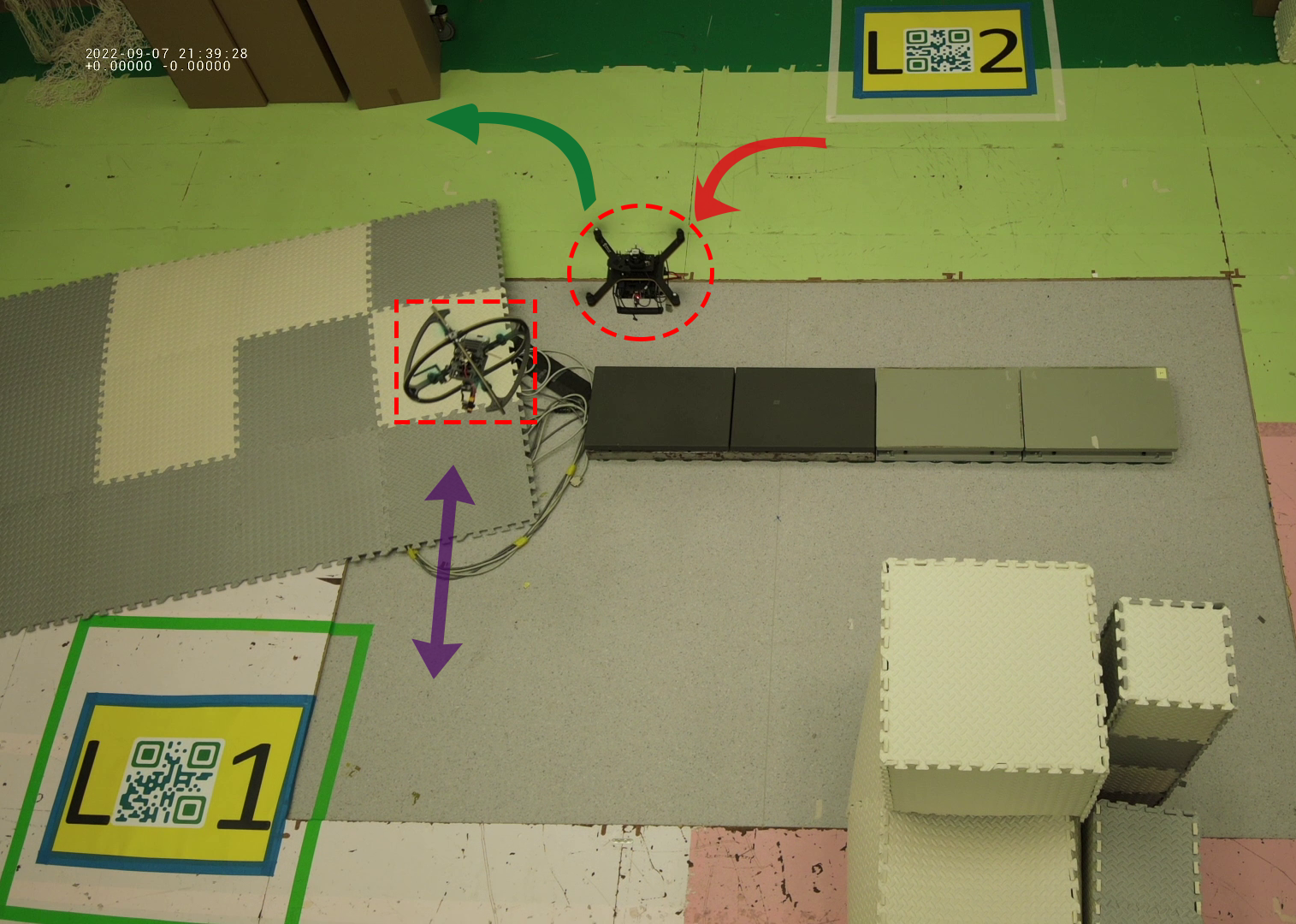} \label{fig:kobs}
}\hfill
		\subfloat[]
{\includegraphics[width=0.22\textwidth]{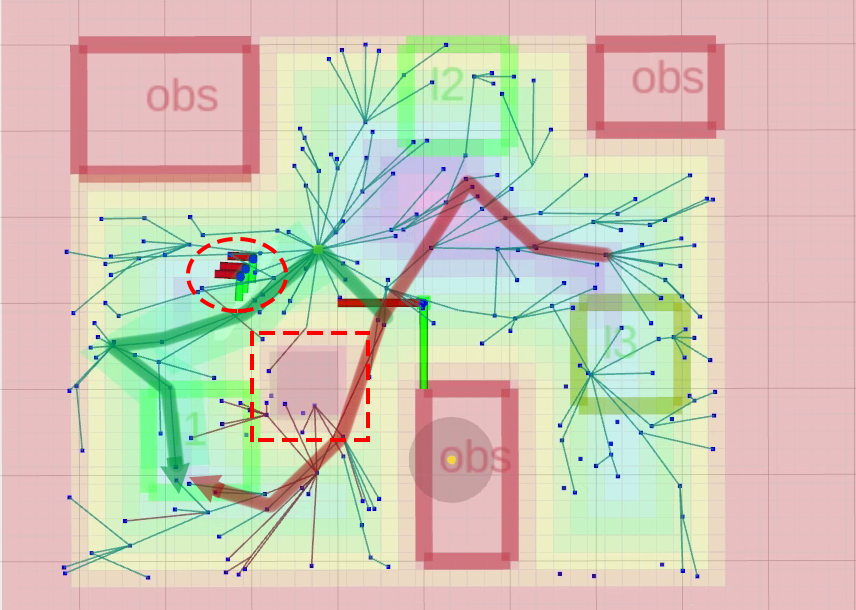} \label{fig:replanObsJT}
}\hfill
	\caption{An indoor experiment of the task $\varphi_2$. (a) The multicopter (in the red dashed circle)  identifies the labeling error in $l_3$ (highlighted in yellow). (b) The old path (dark red) is no longer valid, and a new path (green) takes its place almost instantly. The background is colored according to the distance value in EDT. (c) The multicopter detects an unexcepted obstacle (in the red dashed rectangle) which moves back and forth. (d) A new accepting path is calculated without intersecting the moving obstacle.   }
	\label{fig:indoor_fly}
\end{figure}

One typical execution can be viewed in Fig.~\ref{fig:indoor_fly}. The multicopter takes off at the center of the map and  carries out the initial plan $l_{2}\rightarrow l_{3}\rightarrow l_{2}$. In Fig.~\subref*{fig:kreg}, after getting water, the multicopter flies towards  $l_{3}$  and receives  knowledge $\kappa_{\mathrm{reg}} = (l_{3}, \varnothing)$, indicating that the property of region $l_3$ is incorrect in $X_{\mathrm{init}}$. Benefiting  from the solution library maintained, the multicopter rapidly adopts an alternative solution, turning around and heading to $l_1$ (Fig.~\subref*{fig:replanRegJT}).
In  Fig.~\subref*{fig:kobs}, the multicopter gets knowledge $\kappa_{\mathrm{obs}} = (\upsilon_{1}, (1.36, 0.98, 1.02))$ during flight, and potential collision is detected.
Thanks to the sample reusing strategy, the multicopter quickly alters its course and flies around the obstacle to $l_3$ (Fig.~\subref*{fig:replanObsJT}). 
The multicopter eventually completes the task by  reaching $l_3$ and putting out the fire.
Our planner is shown to successfully accommodate property and occupancy uncertainties through online plan revision and refinement mechanisms.
The full video can be found at https://youtu.be/UH-4KcCUixw.

\section{Conclusion}

In this paper, we have proposed a high-level path planner for mobile robots to complete a temporal logic task regardless of a limited understanding of the workspace. The planner 
quickly obtains a feasible plan and improves it over time during execution.   
We develop two algorithms for node reusing and deduplication to overcome issues that may impair the correctness and effectiveness of our plan-revising mechanism when responding to incidents. 
Results in simulation and real-world missions show that the proposed planner can facilitate online robot tasking in partially-unknown environments.  Future work includes developing a user-friendly interface and  dedicated motion planning algorithms.

\bibliographystyle{IEEEtran}   

\input{root.bbl}

\end{document}